\documentclass[letterpaper, 10 pt, conference]{ieeeconf}  

\IEEEoverridecommandlockouts                              

\usepackage{cite}
\usepackage{amsmath,amssymb,amsfonts}
\usepackage{algorithmic}
\usepackage{graphicx}
\usepackage{textcomp}
\usepackage{cuted}
\usepackage{capt-of}
\usepackage{float}
\usepackage{booktabs}
\usepackage{threeparttable}
\usepackage{mathtools}
\usepackage[mathscr]{eucal}
\usepackage[table,xcdraw]{xcolor}
\usepackage{colortbl}
\usepackage{colortbl}
\usepackage{balance}
\usepackage{color}

\usepackage{marvosym}
\usepackage{algorithmic}
\usepackage{indentfirst}
\usepackage{makecell}

\usepackage{ragged2e}

\makeatletter
\newcommand{\car@semkitfreq}{3.92}
\newcommand{\bicycle@semkitfreq}{0.03}
\newcommand{\motorcycle@semkitfreq}{0.03}
\newcommand{\truck@semkitfreq}{0.16}
\newcommand{\othervehicle@semkitfreq}{0.20}
\newcommand{\person@semkitfreq}{0.07}
\newcommand{\bicyclist@semkitfreq}{0.07}
\newcommand{\motorcyclist@semkitfreq}{0.05}
\newcommand{\road@semkitfreq}{15.30}  %
\newcommand{\parking@semkitfreq}{1.12}
\newcommand{\sidewalk@semkitfreq}{11.13}  %
\newcommand{\otherground@semkitfreq}{0.56}
\newcommand{\building@semkitfreq}{14.1}  %
\newcommand{\fence@semkitfreq}{3.90}
\newcommand{\vegetation@semkitfreq}{39.3}  %
\newcommand{\trunk@semkitfreq}{0.51}
\newcommand{\terrain@semkitfreq}{9.17} %
\newcommand{\pole@semkitfreq}{0.29}
\newcommand{\trafficsign@semkitfreq}{0.08}
\newcommand{\semkitfreq}[1]{{\csname #1@semkitfreq\endcsname}}

\makeatletter
\let\NAT@parse\undefined
\makeatother
\usepackage{hyperref}

\title{\LARGE \bf
Chameleon: Fast-slow Neuro-symbolic Lane Topology Extraction
}

\author{Zongzheng Zhang$^{*1,2}$, Xinrun Li$^{*2}$, Sizhe Zou$^{1}$, Guoxuan Chi$^{1}$, Siqi Li$^{1}$\\
Xuchong Qiu$^{2}$, Guoliang Wang$^{1}$, Guantian Zheng$^{1}$, Leichen Wang$^{2}$, Hang Zhao$^{3}$, and Hao Zhao$^{1}\textsuperscript{\Letter}$
\thanks{$^{1}$Institute for AI Industry Research (AIR), Tsinghua University, China.}
\thanks{$^{2}$Bosch Corporate Research, China.}
\thanks{$^{3}$Institute for Interdisciplinary Information Sciences(IIIS), Tsinghua 
              University, China.}
\thanks{$*$ Equal contribution.}
\thanks{\textsuperscript{\Letter} Corresponding to zhaohao@air.tsinghua.edu.cn}
}

\begin{document}

\maketitle
\thispagestyle{empty}
\pagestyle{empty}

\begin{abstract}
Lane topology extraction involves detecting lanes and traffic elements and determining their relationships, a key perception task for mapless autonomous driving. This task requires complex reasoning, such as determining whether it is possible to turn left into a specific lane. To address this challenge, we introduce neuro-symbolic methods powered by vision-language foundation models (VLMs). Existing approaches have notable limitations: (1) Dense visual prompting with VLMs can achieve strong performance but is costly in terms of both financial resources and carbon footprint, making it impractical for robotics applications. (2) Neuro-symbolic reasoning methods for 3D scene understanding fail to integrate visual inputs when synthesizing programs, making them ineffective in handling complex corner cases. To this end, we propose a fast-slow neuro-symbolic lane topology extraction algorithm, named Chameleon, which alternates between a fast system that directly reasons over detected instances using synthesized programs and a slow system that utilizes a VLM with a chain-of-thought design to handle corner cases. Chameleon leverages the strengths of both approaches, providing an affordable solution while maintaining high performance. We evaluate the method on the OpenLane-V2 dataset, showing consistent improvements across various baseline detectors. Our code, data, and models are publicly available at \href{https://github.com/XR-Lee/neural-symbolic}{https://github.com/XR-Lee/neural-symbolic}

\end{abstract}

\section{Introduction}
Online map perception \cite{jiang2024p} is a crucial topic in modern autonomous driving, with the potential to bypass the need for high-cost high-definition (HD) maps. Current 3D scene understanding methods can effectively detect instances like lanes and traffic elements (e.g., as shown in Fig.~\ref{fig:enter-label}). However, the relationships among these detected instances span a large combinatorial space, requiring extensive labeled data for supervised training. Therefore, we aim to extract lane topology using a few-shot approach leveraging vision-language foundation models (VLMs).

Specifically, we focus on lane topology extraction as defined by OpenLane-V2 \cite{wang2024openlane}, which involves detecting lanes and traffic elements (e.g., traffic lights and signs) and extracting their relationships. This problem is challenging and requires high-level reasoning capabilities, such as determining whether one can drive into a lane given all the traffic elements at an intersection. To our knowledge, existing VLMs are still incapable of directly addressing such complex 3D scene understanding tasks.

As workaround solutions, there are two types of VLM-based scene understanding methods in the literature:

\begin{figure}
    \centering
    \includegraphics[width=1.0 \linewidth]{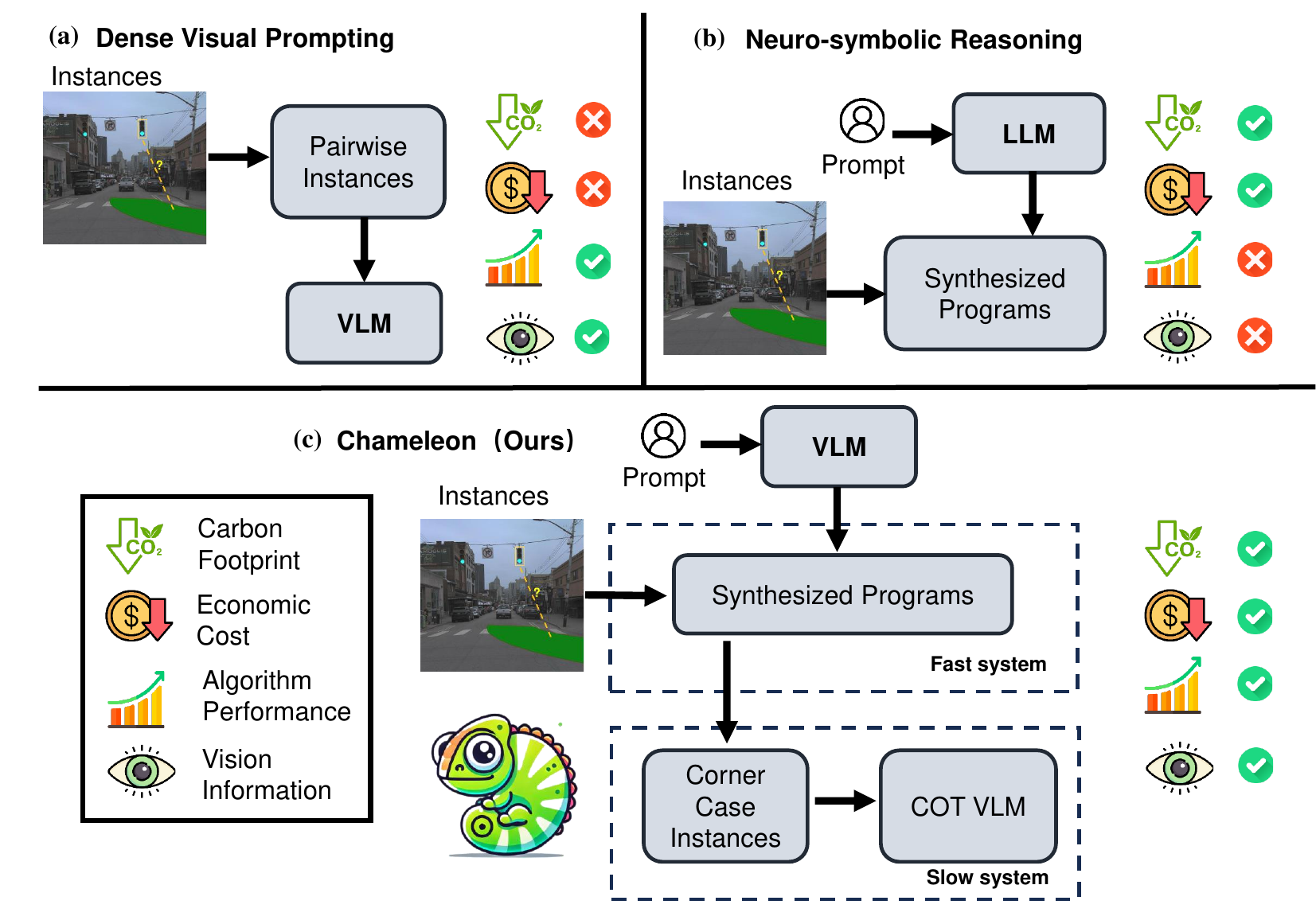}
    \caption{VLMs cannot directly address complex 3D scene understanding tasks, such as lane topology extraction. (a) One possible approach is to use dense visual prompting, as in \cite{shtedritski2023does}, which is accurate but inefficient. (b) Another approach is Neuro-symbolic reasoning, as in \cite{hsu2023ns3d}, which does not effectively leverage visual inputs for program synthesis. (c) Our proposed Chameleon method employs a fast-slow design, where one VLM synthesizes programs and another handles corner cases.}
    \label{fig:enter-label}
    \vspace{-0.8cm}
   
\end{figure}

\begin{figure*}
    \centering
    \setlength{\abovecaptionskip} {5pt} 
    \includegraphics[width=0.88\linewidth]{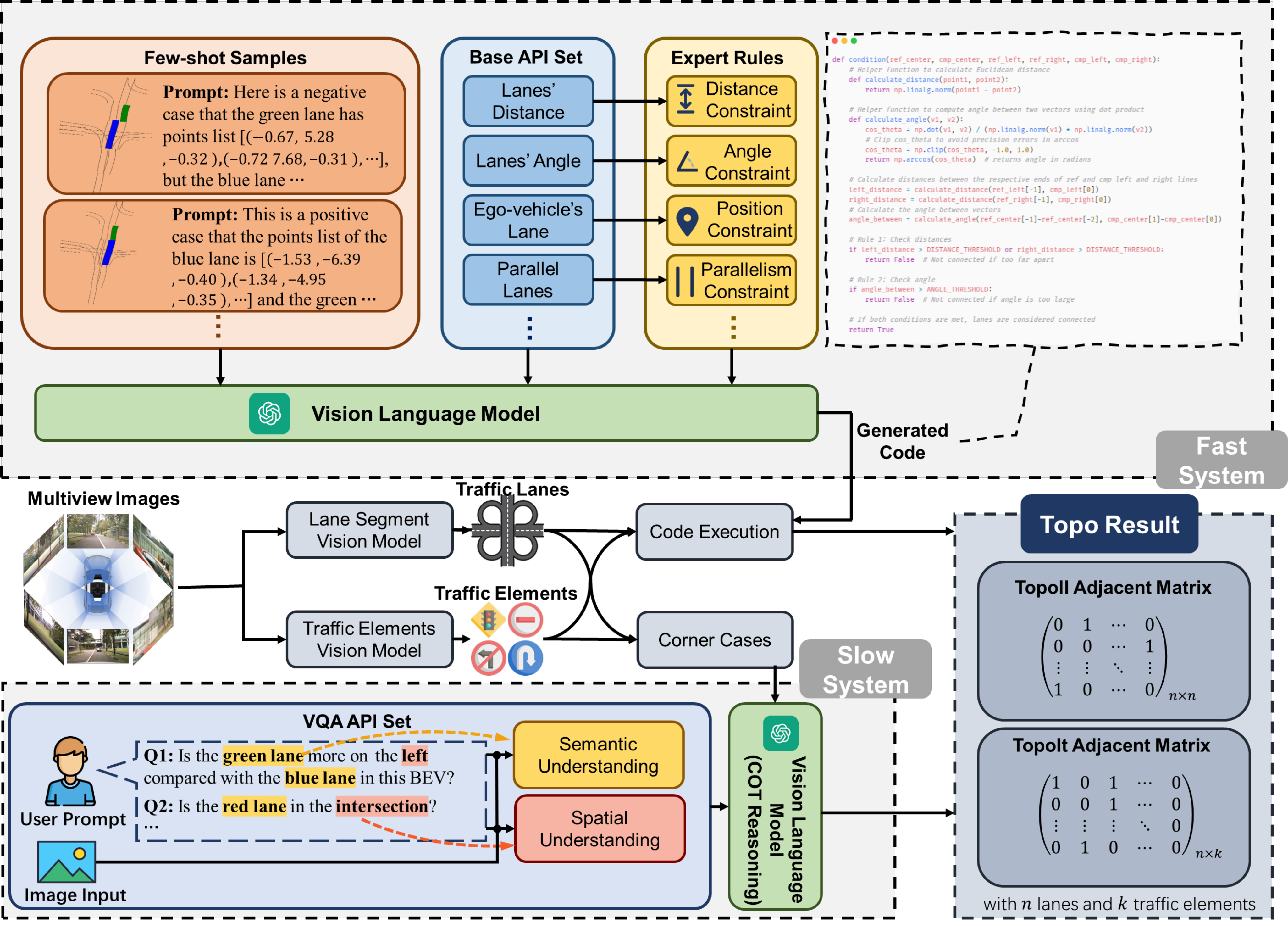}
    \caption{Overview of Chameleon. Given multi-view images as input, the vision models first generate traffic lanes and traffic elements, respectively. The proposed fast system leverages a large Vision Language Model which takes predefined visual-textual few-shot samples and text prompts as inputs, and generates executable codes to process the predictions by the vision models. The proposed slow system consists of a VQA API Set and a Vision Language Model with Chain-of-thought reasoning, where vision prompts and text prompts within the VQA API Set are the inputs of VLM. Subsequently, the topology reasoning results are an combination of code execution results and VLM outputs.}
    \label{fig:main-fig}
    \vspace{-0.6cm}
    
\end{figure*}

(1) Dense visual prompting (Fig.~\ref{fig:enter-label}-a). This paradigm is exemplified by RedCircle \cite{shtedritski2023does}, which places red circles on the image and transforms complex reasoning tasks into question-answering (QA) tasks about the objects within those circles. The red circles act as visual prompts, hence the term ``dense visual prompting''. While lane topology extraction can also be formulated as dense visual prompting, extracting relationship results in an impractically large QA set. For instance, if an image contains 100 lane segments and 20 traffic elements, trivial dense visual prompting would result in $100\times 100 + 20\times 100 = 12,000$ queries to a VLM. This is not only prohibitively expensive in terms of cost and environmental impact but also impractical for real-time robotics applications, due to latency.

(2) Neuro-symbolic reasoning (Fig.~\ref{fig:enter-label}-b). This paradigm is represented by the NS3D method \cite{hsu2023ns3d}, which synthesizes programs using predefined primitive functions based on natural language input and executes them according to visual features extracted from the visual input. While this method only requires one invocation of a language model, it fails to incorporate visual information during the program synthesis, making it ineffective at handling complex corner cases.

In this paper, we combine the strengths of both methods and propose a fast-slow \cite{sinha2024real}\cite{tian2024drivevlm} neuro-symbolic lane topology extractor called Chameleon (Fig.~\ref{fig:enter-label}-c). Unlike conventional neuro-symbolic approaches like NS3D, Chameleon synthesizes programs using a VLM, tailoring the program to the visual input. The system includes a mechanism to identify corner cases that require further reasoning, which are then processed by a VLM using dense visual prompting. As a result, Chameleon dynamically switches between fast and slow systems, balancing efficiency and performance effectively. Furthermore, we present a manually annotated dataset comprising four subtasks, designed specifically for benchmarking performance.

We summarize our three contributions below:

\begin{itemize}
\item We propose Chameleon, the first lane topology extraction method in a few-shot manner that fully leverages both dense visual prompting and neuro-symbolic reasoning. 
\item Chameleon is the first to integrate visual information from VLMs for lane topology extraction, enabling tailored program synthesis for each visual scene.
\item We present a Chain-of-Thought (COT) method to identify corner cases requiring extra reasoning, processing them with dense visual prompting. Chameleon balances computational efficiency and high performance suitable for robotics. Additionally, we establish a comprehensive dataset and benchmark to evaluate the performance of these subtasks.
 
\end{itemize}


\section{Related Works}
\vspace{-5pt}
\textbf{Lane Topology Reasoning.} Structured scene understanding \cite{zhao2017physics,chen2022pq,zhao2020learning,chen2024idea,gao2023semi} is an important computer vision topic with applications in autonomous driving \cite{wu2023mars, zheng2024monoocc, tian2023unsupervised, jin2023adapt}.
Lane topology extraction, the specific problem we attack, can be traced back to DeepRoadMapper \cite{mattyus2017deeproadmapper}, where a segmentation model was combined with the A* search algorithm to retrieve road topology. The lane topology reasoning task was first proposed in OpenLane-V2 \cite{wang2023openlanev2}, where the topological relationships between traffic signs and lanes, as well as relationships between lanes, are represented by an adjacency matrix, with corresponding annotations provided in the dataset. TopoNet \cite{li2023toponet} utilizes a GNN to infer topologies among traffic elements and lanes, while TopoMLP \cite{wu2023topomlp} argues that even a single MLP can be an effective learner for these topologies. LaneGraph2Seq \cite{peng2024lanegraph2seq} reformulates the graph prediction task into a sequence prediction task. TopoLogic \cite{fu2024topologic} enhances interpretability by performing inference based on lane geometric distances and query similarities. A common issue in these works is their heavy reliance on data. Our approach, on the other hand, preserves interpretability in inference while eliminating the need for supervision.

\textbf{Neural Symbolic.}
Neural-symbolic systems leverage the perceptual capabilities of neural networks and the logical reasoning power of symbolic methods, making them essential for applications that require both perception and cognition. NS-3D \cite{hsu2023ns3d} introduces a domain-specific language (DSL) for abstracting operation execution in symbolic systems. Logicode \cite{zhang2024logicode} applies large language models (LLMs) to translate expert knowledge into code for anomaly detection tasks. Rekap \cite{huang2024rekep} employs generated Python scripts to model symbolic constraints in a grasp manipulation system.

Despite these advances, most  neural-symbolic approaches depend on LLMs to generate DSLs or programs, and the potential of visual prompts to enhance symbolic generation remains unexplored. This research addresses this gap by generating programs through VLMs, using few-shot visual samples as prompts.

\begin{figure}
    \centering
    \setlength{\abovecaptionskip}{5pt}
    \setlength{\belowcaptionskip}{-5pt}
    \includegraphics[width=1.0\linewidth]{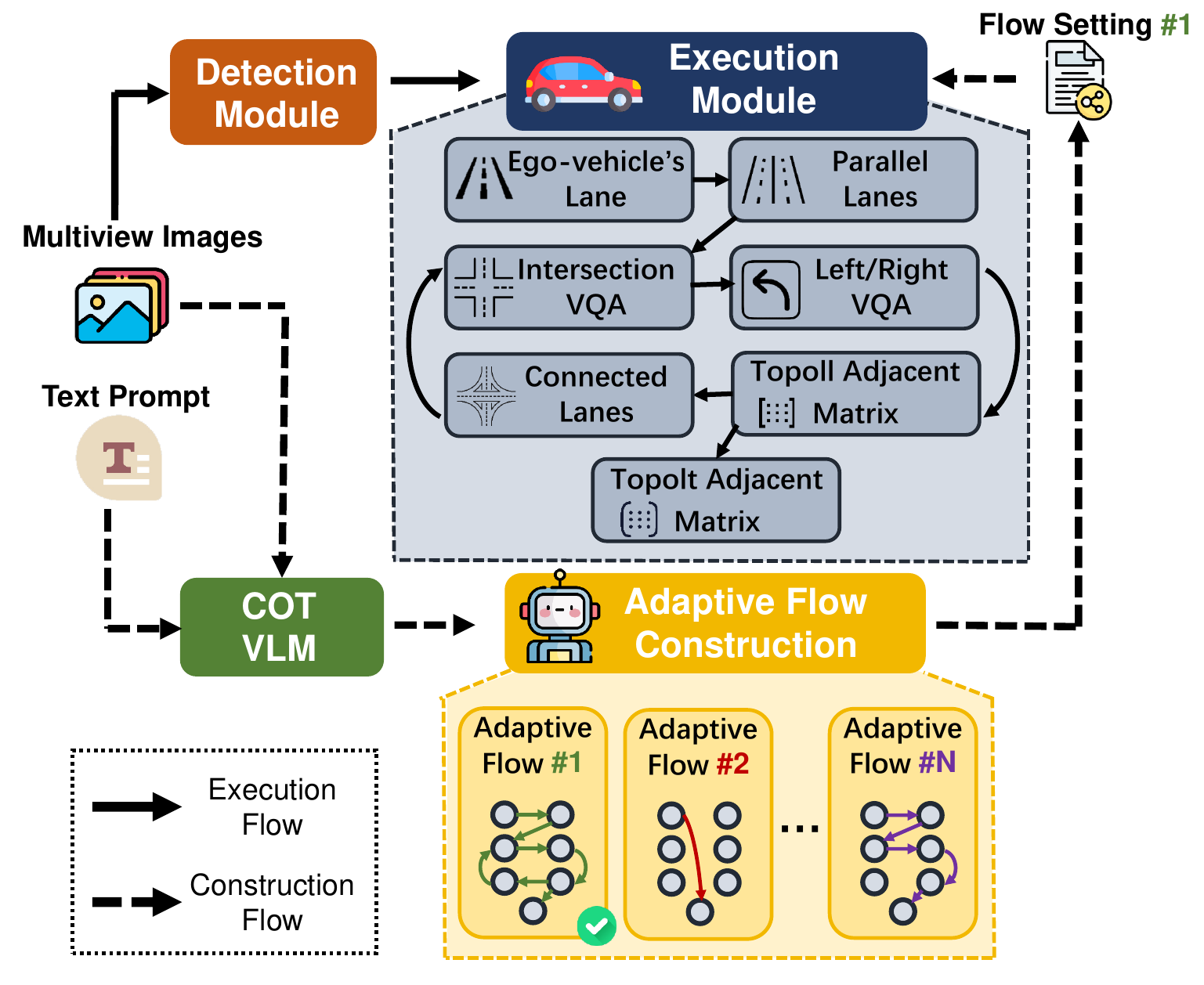}
    \caption{Illustration of Chameleon architecture. Given multi-view images and text prompt as input, Chameleon achieves lane topology extraction. Each API or dense visual prompting VQA task is represented as a node. COT VLM adaptively selects the nodes which are executed to infer the topology results based on input. }
    \label{fig:flow}
    \vspace{-0.6cm}
   
\end{figure}

\textbf{Visual Question Answering.} VQA involves answering natural language questions based on images, requiring models to process and understand both visual and textual data. Early models, like Antol et al.\cite{antol2015vqa}, struggled with fusing these modalities, leading to exploration of more advanced techniques. Inspired by the success of transformers in NLP\cite{vaswani2017attention}, models like VisualBERT~\cite{li2019visualbert} apply transformers to align image regions with text tokens, enabling stronger multimodal representation through joint vision-language pre-training.

Recently, the advancements of LLMs, e.g., GPT-4~\cite{achiam2023gpt4}, LLaMMA~\cite{touvron2023llama}, demonstrate encouraging generalization capacity across tasks such as writing, translation, coding, etc. Built on the success of LLMs, recent VLMs, e.g., GPT-4V~\cite{achiam2023gpt4}, LLaVA~\cite{liu2024llava}, mini-GPT-4~\cite{zhu2023minigpt4}, effectively build the alignment across image-language modalities, and can achieve VQA in a zero-shot manner. In the domain of autonomous driving, Cao et al. recently introduce a large-scale vision-language benchmark MAPLM~\cite{cao2024maplm} for map and traffic scene understanding, and point out that there is still a significant performance gap for road understanding with VLMs. Inspired by existing works, we carefully design our Chameleon framework with VLMs, and achieve competitive few-shot lane topology reasoning performance without any additional model finetuning. 

\section{Method}\label{sec:method}
\vspace{-5pt}
\subsection{\textbf{Overview}}

In lane topology extraction task, we predict dense adjacency matrix, with $A \in \mathbb{R}^{m\times m}$  for lane segment predictions and  $ A \in \mathbb{R}^{m\times n}$,  for traffic element predictions. Here, $m$ and $n$ represent the numbers of lane segments and traffic elements, respectively. 
Although dense visual prompting for question answering can achieve high performance as shown in Tab.~\ref{tab:vqa_metric}, it is prohibitively expensive to only use it for this dense matrix prediction in terms of cost, environmental impact, and speed. Given that the adjacency matrix is mostly sparse, we use a COT method to select corner cases that require dense visual prompting. To improve inference efficiency, we implement a fast-slow system architecture: The fast system, handles basic inference tasks using symbolic representations, while the slow system addresses challenging corner cases and infers the final results. Simple tasks typically involve straight lanes without occluding other vehicles or traffic elements, while corner cases usually occur at intersections with dense traffic and a high number and variety of traffic elements, especially traffic signs.

Fig.~\ref{fig:main-fig} illustrates the overall architecture of the proposed method. Given multi-view images as input, the vision models predict traffic elements and lane segments, respectively. A VLM takes predefined prompts as input and generates executable code to process the predictions efficiently. These prompts include both vision and text prompts: the vision prompts are derived from selected few-shot examples relevant to the tasks, while the text prompts include detailed data from few-shot samples, API descriptions, and expert rules. Subsequently, the system invokes a VLM based on matching conditions, including exceptions, VQA API calls, and corner case summaries.

\begin{table*}[htbp]
\caption{Performance Comparison of Different Methods.
Note that our method sees several topology cases during test time while prior methods are supervised by the topology annotations in the training set of OpenLane-v2\cite{wang2024openlane}.}
\vspace{-3pt}
\centering
\begin{tabular}{cccccc}
\toprule
\textbf{Methods} & \textbf{$\text{DET}_{ls}$(\%)} & \textbf{$\text{DET}_{te}$(\%)} & \textbf{$\text{TOP}_{lsls}$ (\%)} & \textbf{$\text{TOP}_{lste}$(\%)} & \textbf{ } \\
\hline
Lanesegnet\cite{li2023lanesegnet}     & 32.30  & N/A  & 25.40  & N/A & supervised \\
TopoLogic\cite{fu2024topologic}     & 33.0  & N/A  & 30.80  & N/A & supervised \\

MapVision\cite{yang2024mapvision}              & 44.00  & 73.00  & 40.00 & 52.00    & supervised \\
Lanesegnet\cite{li2023lanesegnet}+SMERF\cite{luo2023augmenting}+TopoMLP\cite{wu2023topomlp}    & 38.35  & 54.46  & 31.87                 & 27.32 & supervised \\
Lanesegnet\cite{li2023lanesegnet}+SMERF\cite{luo2023augmenting}+Ours              & 38.35  & 54.46  &\textbf{27.69}  &\textbf{25.73}     & \textbf{zero-shot} \\

Lanesegnet\cite{li2023lanesegnet}+SMERF\cite{luo2023augmenting}+Ours              & 38.35  & 54.46  &\textbf{31.65}  &\textbf{27.97}     & \textbf{3-shot} \\

\midrule
Powerful baseline      & 54.42  & 67.37  &46.59  &N/A     & supervised \\
Powerful baseline + Ours          & 54.42  & 67.37      &\textbf{34.16}     &\textbf{45.74}  & \textbf{zero-shot} \\
Powerful baseline + Ours          & 54.42  & 67.37      &\textbf{46.54}     &\textbf{53.05}  & \textbf{3-shot} \\


\bottomrule 
\end{tabular}
\label{tab:zeroshot}
\vspace{-0.6cm}
\end{table*}

\vspace{-5pt}
\subsection{\textbf{Prompts}}
To perform symbolic reasoning, we use various prompts to generate symbolic codes. These include visual prompts with few-shot references to positive or negative cases, API descriptions, and expert rules.

\textbf{API Prompts.} The API prompts define the input and output for the generated code, as well as the API input-output descriptions such as function for self-localization on lane and parallel-lane search, etc. In our implementation, we also define selected VQA tasks as APIs during program synthesis.

\textbf{Expert Rules Prompts.}
To stabilize the code generation process and incorporate prior knowledge from domain experts, we add expert rules as prompts for program synthesis. In ${TOP}_{lsls}$ task, angle and distance constraints are enforced. For example, the end point of the parent lane should not be too far from the start point of the child lane in order to satisfy driving geometry constraints. The ${TOP}_{lste}$ utilized rules which specify no lane topology is permitted inside intersection.

\textbf{Few-shot Prompts.} 
In a few-shot scenario, we select positive and negative examples and render them in the camera's perspective view. We also convert the coordinates of these examples into text, serving as visual prompts and textual prompts, respectively.

\textbf{VQA Prompts.} 
For the VQA tasks, the text prompt consists of simple questions regarding semantic and spatial contexts. We also utilize a COT prompt. The visual prompts are rendered images based on predictions from both the perspective view and the bird's-eye view.

\subsection{\textbf{Code Execution}}
For the generated programs, the code execution processes differ between the ${TOP}_{lsls}$ and ${TOP}_{lste}$ tasks. For the ${TOP}_{lsls}$ task, a simple pairwise pre-defined code framework is used, where the VLM generates Python code based on the API description and given prompts. This code, produced as a string, is then executed using Python's $exec$ function. In contrast, the ${TOP}_{lste}$ task involves more API calls(Fig.~\ref{fig:flow}), so we utilize OpenAI’s function-calling API to manage the required function executions. Firstly, we prompt the VLM to generate a Chain-of-Thought that resolves the topology extraction problem, which has six steps as shown in Fig.~\ref{fig:flow} Execution Module. This is further used as text prompt for VLMs to synthesize programs that may skip certain steps in the chain adaptively according to visual inputs. Some of steps involve corner cases that need to be addressed by a dense visual prompting VLM model thus sent into the slow system. By summarizing the API results, the system can infer a potential topology pair.


\subsection{\textbf{Dense Visual Prompting VQA Tasks}}

Dense visual prompting VQA tasks act as the core APIs for our slow-system, particularly for interoperable processes in open scene topology inference. To test the capabilities of VLM models, we created several basic VQA tasks.
As illustrated in Tab.~\ref{tab:vqa_metric}, we focus on four different tasks. In the Left-or-Right task, two lane segments are presented in bird-eye-view (BEV).  The model is required to perform a three-class classification, choosing from left, right, or no relationship. For the Is-in-Intersection task, a single lane is shown in both perspective and BEV views.  The model must determine if the lane is inside an intersection. In the adjacency task, two lanes are given, and the model has to judge whether they are adjacent to one another. Finally, in the Vector task, the model must assess if the directions of two rendered vector arrows are similar.

\section{Experiments}

\begin{table*}[htbp]
\caption{Dense Visual Prompting Examples. Note that this paradigm is inspired by the RedCircle method \cite{shtedritski2023does}. But trivially doing this kind of dense visual prompting for lane topology is not practical, so we only use this in the slow system.}
\vspace{-5pt}
\centering
\begin{tabular}{>{\centering\arraybackslash}m{0.05\textwidth} >{\centering\arraybackslash}m{0.14\textwidth} >{\centering\arraybackslash}m{0.50\textwidth} >{\centering\arraybackslash}m{0.16\textwidth}}
\toprule
\textbf{Task} & \textbf{Input Image} & \textbf{Prompt} & \textbf{Output}   \\
\midrule

Left-or-Right     &  \includegraphics[width=0.15\textwidth]{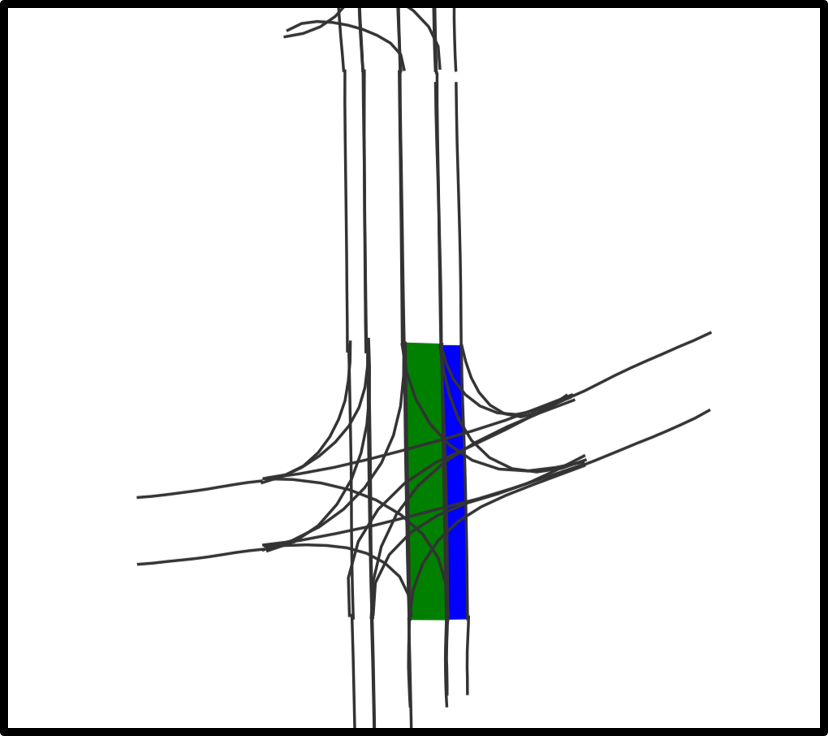}   & \noindent\justifying{\texttt{\textbf{Description:} In the provided bird's-eye view (BEV), the black lines in the photos are lane boundaries that are only for references. Color blocks highlighted are different segments of lanes. 
The colors of the blocks come from green and blue.}}

\noindent\justifying{\texttt{\textbf{Question:} You are an expert in determining positional relationships of lane segments in the image. Is the green segment on the left of the blue segment, or on the right? Please reply in a brief sentence.}}    & \noindent{\justifying{\texttt{The green segment is on the left of the blue segment.}}}   \\
\midrule

Is-in-Intersection              &  \includegraphics[width=0.15\textwidth]{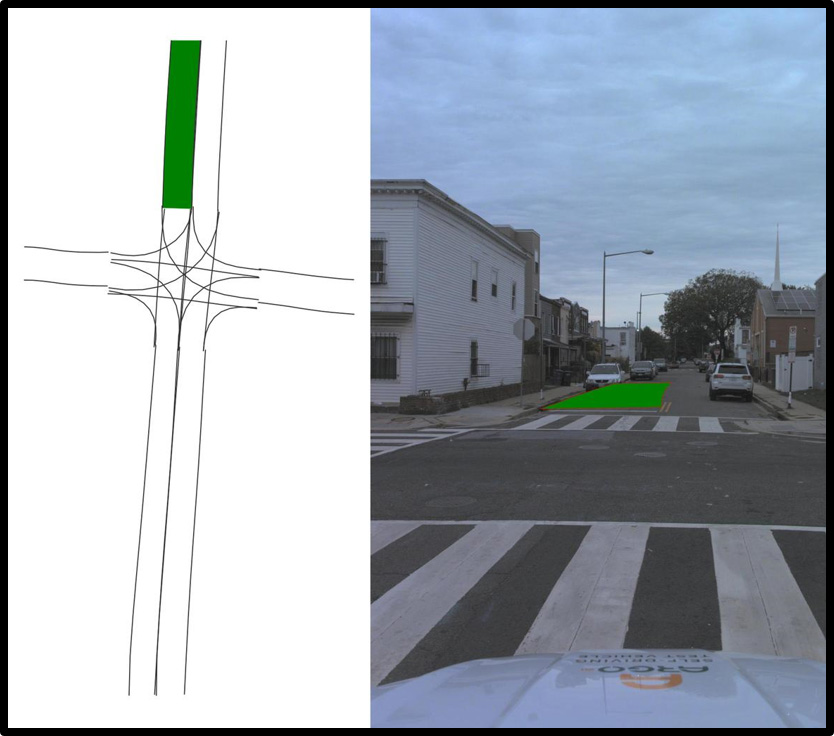}   & \noindent\justifying{\texttt{\textbf{Description:} The provided photo is mosaiced with two images, with the bird's-eye view (BEV) on the left and the front perspective view (PV) on the right.
Normally, the lane segment is considered as not in the intersection when it is in front of the area or at rear of the area.}}

\noindent\justifying{\texttt{\textbf{Question:} You are an expert in extracting the positional information of lane segments. Let's determine if the the green segment patch is in the intersection area. Please reply in a brief sentence starting with "Yes" or "No".}}    &  \noindent{\justifying{\texttt{Yes, the green segment patch is in the intersection area.}}} \\
\midrule

Adjacency  &  \includegraphics[width=0.15\textwidth]{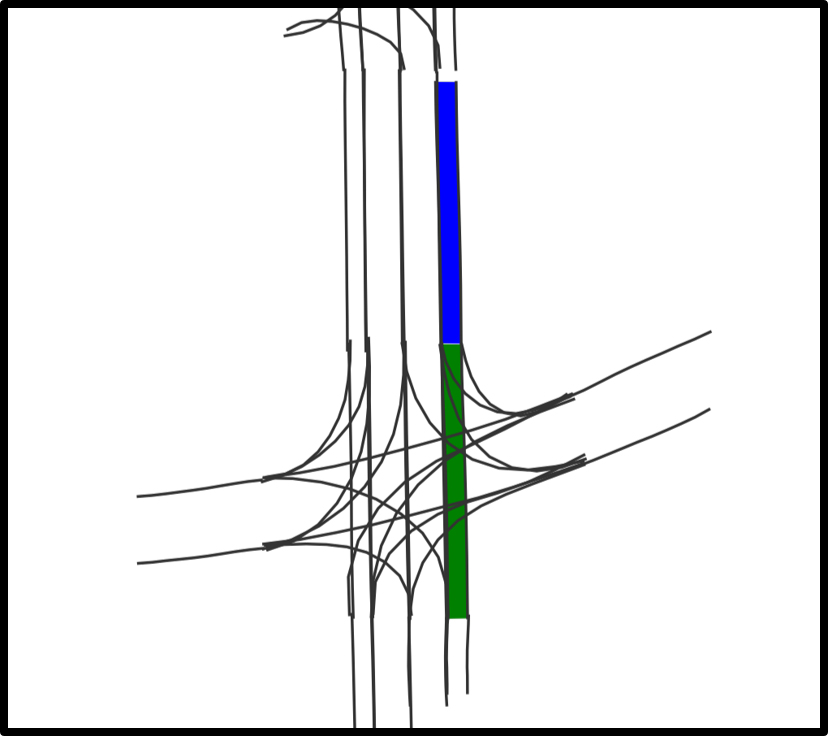}   & \noindent\justifying{\texttt{\textbf{Description:} In the provided bird's-eye view (BEV), the black lines in the photos are lane boundaries. Color blocks highlighted are different segments of lanes. Only two lane segments in the same lane end to end adjacent are considered as directly connected.}}

\noindent\justifying{\texttt{\textbf{Question:} You are an expert in determining adjacency of lane segments. Let's determine if the the green patch is directly connected with the blue patch. Please reply in a brief sentence starting with "Yes" or "No".}}    &  \noindent{\justifying{\texttt{Yes, the green patch is directly connected with the blue patch.}}} \\
\midrule

Vector    &  \includegraphics[width=0.15\textwidth]{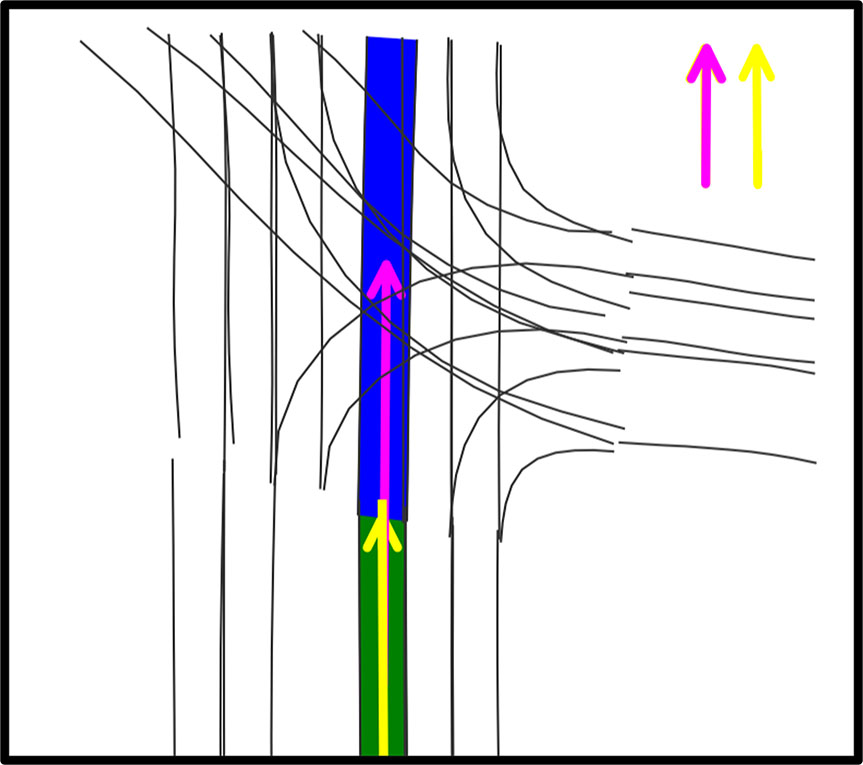}   & \noindent\justifying{\texttt{\textbf{Description:} In the provided bird's-eye view (BEV), the green and blue lane segments are highlighted. The arrow represents the directions of lanes. Normally or when the case is confusing, two directions with deviation of less than 45 degrees are considered as compatible. }}

\noindent\justifying{\texttt{\textbf{Question:} You are an expert in determining direction relationships of lane segments. Let's determine if the directions of the two arrows match. Please reply in a brief sentence starting with "Yes" or "No".}}    &  \noindent{\justifying{\texttt{Yes, the directions of the two arrows match.}}}  \\
\bottomrule
\end{tabular}
\label{tab:vqa_metric}
\end{table*}

\begin{table*}[htbp]
\caption{VQA Performance Comparison of Different Methods on downsampled OpenLane-V2  \cite{wang2024openlane} validation set}
\vspace{-5pt}
\centering
\begin{tabular}{cccccc}
\toprule
\textbf{Method} & \textbf{Left-or-Right (\%)} & \textbf{Is-in-Intersection (\%)} & \textbf{Adjacency (\%)} & \textbf{Vector (\%)}  & \textbf{Average (\%)}\\
\hline
MLP (Supervised)      &  90.20   & 72.00    & 97.06    & 98.04  & 89.33\\
GPT-4o                &  82.43   & 77.59    & 88.65   & 81.30 & 82.49      \\
GPT-4-vision-preview  &  67.49   & 53.43    & 70.00    & 78.91 & 67.46         \\
LLaVA 1.5-13b        &  53.60   & 58.72    & 57.39     & 50.57 & 55.07         \\
\hline
\end{tabular}
\label{tab:vqa}
\vspace{-0.6cm}
\end{table*}

\begin{figure*}
\centering
  \includegraphics[width=0.85\textwidth]{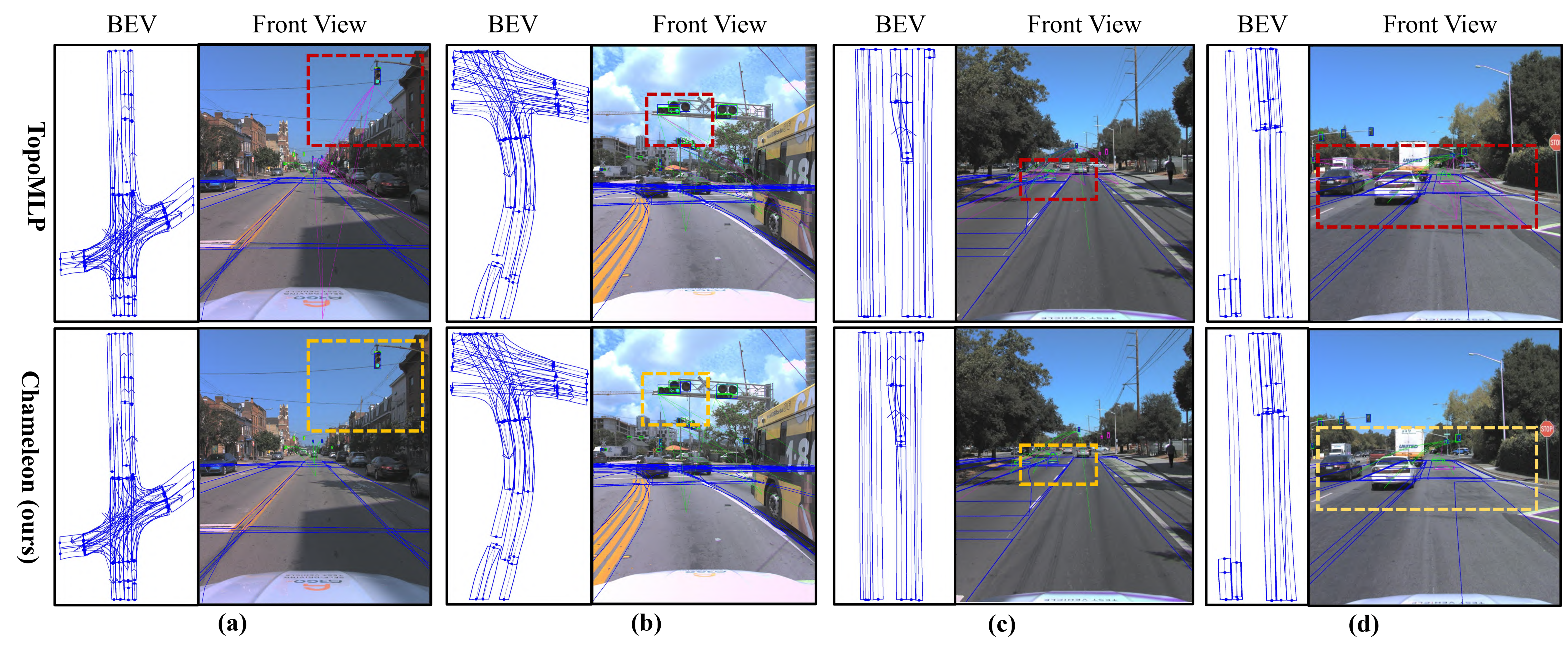}
  \setlength{\abovecaptionskip}{-10pt}
  \caption{Qualitative results of TopoMLP and Chameleon(ours) on Openlane-V2 \cite{wang2024openlane} validation dataset. (a)~The vehicle has just passed the intersection. (b)~There is a left-turn traffic light ahead. (c)~The ground lane is marked with a straight-ahead sign. (d)~The vehicle is on a one-way right-turn lane. The selected scenes are all corner cases and have undergo further reasoning through dense visual prompting.}
 \vspace{-0.6cm}
  \label{fig:qualitative_exp_fig}
\end{figure*}

\subsection{\textbf{Experiment Settings}}

\textbf{Datasets.} We evaluate our methods on the Openlane-V2 official validation dataset which provides topology annotations between lane segments and traffic elements. 

In addition, we manually annotate a dataset of 500 samples containing four dense visual prompting VQA subtasks, which is used to assess the performance metrics of each task. This dataset remains broadly applicable to other autonomous driving scenarios. 

\textbf{Evaluation metrics.} Following common practices, we report the Mean Average Precision (mAP) for the tasks of lane segments detection and traffic elements detection. For the topology task, we use the official metric $\text{TOP}_{lsls}$ for mAP on the topology among lane segments, and $\text{TOP}_{lste}$ for mAP on the topology between lane segments and traffic elements. For the VQA classification problem, since positive and negative samples are balanced during annotation, we use Accuracy as the metric to evaluate VQA performance.

\subsection{\textbf{Implementation Details}}\label{details}

In this study, we apply our approach to self-implemented baseline and conducted experiments. Specifically, we leverage the standard definition map (SD) encoding and fusion module from SMERF \cite{luo2023augmenting} and the LanesegNet detection framework \cite{li2023lanesegnet} to perform lane segment detection. The 2D traffic element detection is handled using DETR \cite{zhu2020deformable}, while TopoMLP is trained concurrently to predict topological relationships. 
In the more powerful baseline named ``powerful baseline'' in Tab.~\ref{tab:zeroshot}, we enhanced the first baseline by incorporating temporal information based on streammapnet\cite{Yuan_2024_streammapnet}, replacing the backbone with a larger one using Vovnet, and using YOLOv8 based on the training scheme from \cite{wu20231st} to detect traffic elements. For the second baseline, two vision models are trained separately using extensive data, we only provided the TopoMLP baseline for lsls. 

For the GPT-4 models, we use the official model API, GPT-4-vision-preview (Access Date: November 6, 2023) and GPT-4o API (Access Date: May 13, 2024). For LLaVA, we implement the official weights LLaVa-v1.5-13b-full\_ft-1e.

In Tab.~\ref{tab:zeroshot}, refer to scenarios with a limited number of training samples in the input prompt. In our few-shot learning case, we use a 3-shot configuration, meaning three frames and their corresponding annotations are included as references.

For the VQA benchmark, the MLP method is based on a classification model using ResNet18 as the backbone. The dataset is split into a 3:1 ratio for training and testing. The model is trained with the Adam optimizer and cross-entropy loss over 20 epochs.

\vspace{-5pt}
\subsection{\textbf{Quantitative and Qualitative Results}}

\textbf{Comparison with state-of-the-art methods.} In this section, we compare the proposed few-shot method with state-of-the-art supervised methods on the validation set of Openlane-V2. Tab.~\ref{tab:zeroshot} shows comparison results with the methods LaneSegNet, TopoLogic, and MapVision.

Our method is tested using two different baselines, each utilizing a different backbone. The baselines are implemented based on LaneSegNet and TopoMLP with SD encoding and a fusion process from SMERF. As shown in the table, our method achieves competitive performance in the few-shot setting. In the other word, we get similar results to the supervised baseline and even slightly outperform it on the ${TOP}_{lste}$ task, even when compared to fully supervised models. Overall, our method demonstrates significant competitive performance compared to other baselines with only a few-shot manner.

\textbf{VQA Comparison with different VLM methods.} Due to its generality, VQA is compatible with various VLMs. Tab.~\ref{tab:vqa_metric} compares the performance of different VLMs across four tasks. On these tasks, we find that GPT-4 achieved performance comparable to supervised classifier models, while LLaVA performed poorly in both semantic and spatial understanding tasks.
\vspace{-10pt}

\begin{table}[ht]
\centering
\caption{Data and Inference Efficiency comparision}
\vspace{-5pt}
\begin{tabular}{ccc}
\toprule
\textbf{Method}  
&\textbf{Data  } 
&\textbf{Inference Latency} \\
\midrule

Supervised Model       & Full annotation        & 0.1 $\sim$\ 0.7s   \\
Dense Visual Prompting & Zero-shot Samples      & $>$ 200s   \\
Chameleon(Ours)        & 3-shot Samples       & 0.1 $\sim$\ 8s     \\

\bottomrule
\end{tabular}
\label{tab:efficiency}
\vspace{-0.3cm}
\end{table}

To compare the inference costs between different methods, we tested the average VQA task latency on LLaVA using an RTX 4080 GPU. The average VQA latency is 1,447ms, with 6 executions per frame, resulting in 8.7s per frame for the slow system. TopoMLP latency varies from 140ms to 700ms depending on the backbone and resolution. Dense visual prompting, based on 1-by-1 VQA from a 20×20 matrix, results in over 200s per frame. The detail is summarized in Tab.~\ref{tab:efficiency}.

\textbf{Qualitative Results.} To demonstrate the performance of our algorithm more intuitively, we also provide the qualitative visualization of the predicted ls-ls relationship and ls-te relationship in Fig.~\ref{fig:qualitative_exp_fig}. All the compared scenes are corner cases. Each subfigure consists of both BEV and front view. The blue line represents the lane segment detection results, the green line represents ls-te true positives, and the pink line represents ls-te false positives. When the vehicle has just passed the intersection (Fig.~\ref{fig:qualitative_exp_fig}-a), the green light directly above the vehicle does not have a topological relationship with the lane ahead of the intersection. Chameleon (ours) comprehends the spatial relationship between the green light and lanes, thereby making correct judgments, whereas TopoMLP does the opposite.
In Fig.~\ref{fig:qualitative_exp_fig}-b, the left-turn traffic light is topologically linked only to the leftmost lane. Unlike TopoMLP, Chameleon correctly disregards any relationship with the right lanes. The ground lane is marked with a straight-based sign (Fig.~\ref{fig:qualitative_exp_fig}-c), so the sign is only related to its own lane and connected lanes instead of other parallel lanes. Chameleon does it but TopoMLP does not. The vehicle is on a one-way right-turn lane (Fig.~\ref{fig:qualitative_exp_fig}-d) where the green lights controlling straight traffic on both sides don't affect the vehicle.  
Ours solution makes the correct determination, not identifying a topological relationship between the green lights and the lane.

\subsection{\textbf{Ablation Study}}

We conduct ablation studies for ${TOP}_{lsls}$ task on the Openlane-V2 validation set to evaluate the effectiveness of each component in our framework, with results shown in Tab.~\ref{tab:ablation_study_lsls}. The `Prompt to Code' refers to the basic neural-symbolic reasoning with only API prompts. Due to the code instability, we report the average of three symbolic reasoning results as the final performance. The `Expert Rule' refers to incorporating expert observations into the prompt. For the few-shot examples, we introduce three positive and negative cases to improve the generated program.
\vspace{-0.3cm}

\begin{table}[ht]
\centering
\caption{Ablation Study of Prompt Components}
\vspace{-5pt}
\begin{tabular}{ccc}
\toprule
\textbf{Method}  & \textbf{ $\text{TOP}_{lsls}$ (\%)}     \\
\midrule
Pair-wise Prompt to symbolic     & 24.35         \\
+ Expert Rule                    & 27.69         \\
+ Few-shot examples              & 29.23          \\
        \bottomrule
    \end{tabular}
    \label{tab:ablation_study_lsls}
\vspace{-0.3cm}
\end{table}
\vspace{-0.1cm}
\section{Conclusion}

This paper introduces Chameleon, a novel approach that combines dense visual prompting with neuro-symbolic reasoning to extract lane topology using VLMs in a few-shot manner. Chameleon synthesizes programs with visual information, adapts to specific scenes, and handles corner cases efficiently. By balancing computational efficiency with high performance, Chameleon is suitable for real-time robotics applications and showcases the potential of integrating visual inputs into program synthesis for complex 3D tasks.


\balance

\bibliographystyle{IEEEtran}
\bibliography{ref}
\end{document}